\documentclass[conference]{IEEEtran}
\IEEEoverridecommandlockouts

\usepackage{cite}
\usepackage{amsmath,amssymb,amsfonts}
\usepackage{algorithmic}
\usepackage{graphicx}
\usepackage{textcomp}
\usepackage{xcolor}
\usepackage{subcaption}
\usepackage{hyperref}
\usepackage{makecell}
\usepackage{color, colortbl}
\usepackage{xcolor}
\def\BibTeX{{\rm B\kern-.05em{\sc i\kern-.025em b}\kern-.08em
    T\kern-.1667em\lower.7ex\hbox{E}\kern-.125emX}}

\definecolor{LightBlue}{RGB}{212, 250, 252} 

\usepackage{tikz}
\newcommand\copyrighttext{%
  \footnotesize \textcopyright 2024 IEEE.  Personal use of this material is permitted.  Permission from IEEE must be obtained for all other uses, in any current or future media, including reprinting/republishing this material for advertising or promotional purposes, creating new collective works, for resale or redistribution to servers or lists, or reuse of any copyrighted component of this work in other works.}
\newcommand\copyrightnotice{%
\begin{tikzpicture}[remember picture,overlay]
\node[anchor=south,yshift=10pt] at (current page.south) {\fbox{\parbox{\dimexpr\textwidth-\fboxsep-\fboxrule\relax}{\copyrighttext}}};
\end{tikzpicture}%
}

\begin{document}

\title{Evaluating Adversarial Attacks on Traffic Sign Classifiers beyond Standard Baselines}

\author{\IEEEauthorblockN{Svetlana Pavlitska$^{1,2}$, Leopold Müller$^{2}$, J. Marius Z\"ollner$^{1,2}$}
\IEEEauthorblockA{$^{1}$\textit{FZI Research  Center  for  Information  Technology}\\
$^{2}$\textit{Karlsruhe Institute of Technology (KIT)}\\
Karlsruhe, Germany \\
pavlitska@fzi.de}
}

\maketitle

\copyrightnotice
\thispagestyle{empty}
\pagestyle{empty}

\begin{abstract}
  Adversarial attacks on traffic sign classification models were among the first successfully tried in the real world. Since then, the research in this area has been mainly restricted to repeating baseline models, such as LISA-CNN or GTSRB-CNN, and similar experiment settings, including white and black patches on traffic signs. In this work, we decouple model architectures from the datasets and evaluate on further generic models to make a fair comparison. Furthermore, we compare two attack settings, inconspicuous and visible, which are usually regarded without direct comparison. Our results show that standard baselines like LISA-CNN or GTSRB-CNN are significantly more susceptible than the generic ones. We, therefore, suggest evaluating new attacks on a broader spectrum of baselines in the future. Our code is available at \url{https://github.com/KASTEL-MobilityLab/attacks-on-traffic-sign-recognition/}.
\end{abstract}


\section{Introduction}\label{introduction}

Autonomous vehicles and intelligent transportation systems have brought a paradigm shift in global mobility and traffic management. A crucial technology enabling this shift is traffic sign recognition (TSR), a subset of computer vision responsible for identifying and interpreting traffic signs~\cite{stallkamp2011gtsrb}. These TSR systems are indispensable in bolstering road safety and facilitating efficient vehicle navigation~\cite{mammeri2013design}. Despite achieving remarkable performance levels, systems based on deep neural networks exhibit notable vulnerabilities~\cite{houben2022inspect}, particularly their susceptibility to adversarial attacks has ignited considerable concern~\cite{eykholt2018robust,yang2021targeted,zhong2022shadows,woitschek2021physical,li2021adaptive,nuding2020poisoning,khan2022hybrid,pavlitskaya2022adversarial}.

Adversarial attacks are intentional manipulations of input data designed to mislead deep learning models~\cite{goodfellow2014explaining,szegedy2013intriguing}. These attacks can be performed in the real world by placing a print-out with perturbations directly to the scene\cite{brown2017adversarial,thys2019fooling,pavlitskaya2020feasibility,pavlitskaya2022suppress,pavlitskaya2022feasibility}. In our previous work~\cite{pavlitska2023adversarial}, we provide a comprehensive overview of the existing attacks on traffic sign classifiers and detectors. This overview has shown that the research on adversarial attacks on traffic sign classification models has been restricted to repeating baseline models and experiment settings. Even new methods~\cite{chi2023public,hsiao2024natural} are usually demonstrated on \texttt{LISA-CNN} and \texttt{GTSRB-CNN} architectures~\cite{eykholt2018robust}, not allowing for fair comparison across models and datasets. On the other hand, some works use only generic image classification models~\cite{wang2024traffic}, ignoring the established baselines.

In this work, we aim to close two gaps present in the existing works: (1) we decouple model architectures from the datasets they are usually trained on and also compare them to generic models, and (2) we evaluate and compare two attack settings: inconspicuous and visible, which are usually regarded separately without direct comparison~\cite{eykholt2018robust,lu2017adversarial}. To ensure that attacks work beyond the used data, we apply them in a universal manner~\cite{moosavi2017universal}.  To the best of our knowledge, our work is the first to evaluate established baselines beyond the datasets they were trained on. It is also the first to compare these established baselines to the generic CNNs on clean and adversarially perturbed data. 

\begin{figure}[t]
    \centering
    \begin{tabular}{c c c }
        \includegraphics[width=0.27\linewidth]{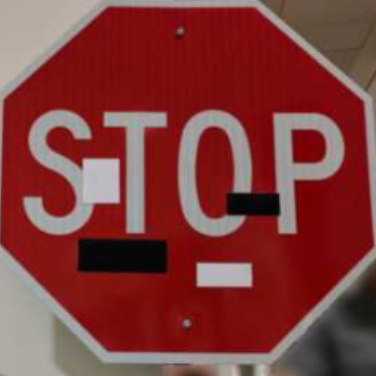} & \includegraphics[width=0.27\linewidth]{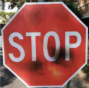} & \includegraphics[width=0.27\linewidth]{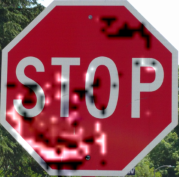} \\
        Eykholt et al.~\cite{eykholt2018robust} & Yang et al.~\cite{yang2021targeted} & Ours\\
        \multicolumn{3}{c}{a) Inconspicuous adversarial attacks} \\
        \includegraphics[width=0.27\linewidth]{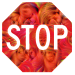} & \includegraphics[width=0.27\linewidth]{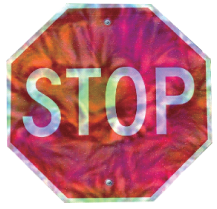} & \includegraphics[width=0.27\linewidth]{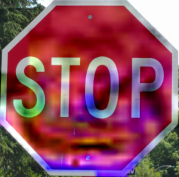}\\
        Chen et al.~\cite{chen2018shapeshifter} & Lu et al.~\cite{lu2017adversarial} & Ours \\
        \multicolumn{3}{c}{b) Visible adversarial attacks} \\
    \end{tabular}
    \caption{Inconspicuous (a) and visible (b) adversarial perturbations of a stop sign.}
    \label{fig:stopsignattacks}
\end{figure}

\section{Related work}\label{relatedwork}

Traffic sign recognition (TSR) involves (1) the localization of a traffic sign, a process known as traffic sign detection (TSD), and (2) the identification of a specific type of sign detected, referred to as traffic sign classification (TSC).

Datasets like the GTSRB~\cite{stallkamp2011gtsrb} or LISA~\cite{mogelmose2012vision} typically include only a class label for each image, thus making them suitable solely for TSC tasks. Among the numerous models designed for TSR tasks, some repeatedly serve as baselines for adversarial attacks. These are convolutional neural networks (CNNs), including the LISA-CNN~\cite{eykholt2018robust}, GTSRB-CNN~\cite{eykholt2018robust}, and a CNN with a spatial transformer~\cite{Jaderberg2015spatial,garcia2018deep}. Despite achieving excellent performance on their respective datasets, comparison of these models presents a challenge due to the varied nature of the datasets. This also includes attacks that target different model baselines. To address this, our methodology will employ both datasets and all three models as baselines for adversarial attacks, enabling more consistent comparisons.

Our previous work~\cite{pavlitska2023adversarial} presenting an overview of existing attacks on TSR shows that while this area appears to be rich in research, on closer inspection, only a limited number of realistic white-box attacks specifically target traffic sign classification. In particular, Eykholt et al.~\cite{eykholt2018robust} developed an algorithm called \textit{Robust Physical Perturbations (RP2)} that creates adversarial examples capable of misleading deep learning models in the physical world. The approach combines physical and synthetic transformations to model environmental conditions, creating effective adversarial perturbations under different real-world conditions.

Conversely, Yang et al. introduced a novel method for real-world road sign recognition called the \textit{Targeted Attention Attack (TAA)}~\cite{yang2021targeted}. This method uses soft attention maps to emphasize crucial pixels while ignoring non-contributing areas. The TAA optimizes a single perturbation or noise based on a set of training images guided by a pre-trained attention map, providing advantages such as transferability and generalization. Compared to the RP2, the TAA has improved the attack success rate and reduced the perturbation loss.

Since then, several new attacks, especially in black-box settings~\cite{woitschek2021physical}, have been proposed. However, even the most recent works still use the standard LISA-CNN and GTSRB-CNN~\cite{chi2023public,zhong2022shadows,hsiao2024natural}. Several other  works~\cite{ye2021patch,wang2024traffic} evaluate only generic image classification models like ResNet~\cite{he2016deep} or AlexNet~\cite{krizhevsky2012alexnet}. Our work directly compares adversarial attacks on established baselines like LISA-CNN and GTSRB-CNN and the generic image classification models.

\section{Attack Approach}

In the following section, we describe our attack strategy. Similar to Eykholt et al.~\cite{eykholt2018robust} and Yang et al.~\cite{yang2021targeted}, we use a two-stage approach, which involves (1) mask creation and (2) attack generation. While the resulting perturbations in these attacks are particularly subtle, resembling graffiti or dirt, more blatant manipulations of traffic signs also exist. To investigate a possible trade-off between the effectiveness of the attack and its conspicuousness, we also evaluate a variant similar to that by Chen et al.~\cite{chen2018shapeshifter} and Lu et al.~\cite{lu2017adversarial} (see Figure \ref{fig:stopsignattacks}). These attacks focus on detection, so the only similarity to our approach is the optical result.

\subsection{Phase I - Mask Creation}
This phase aims to find vulnerable regions of the sign we want to attack. In~\cite{yang2021targeted}, this is done with an attention map; in~\cite {eykholt2018robust}, they follow a similar procedure. We take an image of a traffic sign $I \in \mathbb{R}^{H \times W \times C}$ of class $c$ and resize it to the input dimension of the model $\Theta$. 

\begin{figure}[t]
    \centering
    \begin{subfigure}[b]{0.24\columnwidth}
        \centering
         \includegraphics[width=\linewidth]{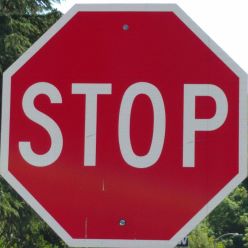}
    \end{subfigure}
    \begin{subfigure}[b]{0.24\columnwidth}
        \centering
         \includegraphics[width=\linewidth]{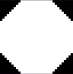}
    \end{subfigure}
    \begin{subfigure}[b]{0.24\columnwidth}
        \centering
         \includegraphics[width=\linewidth]{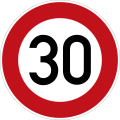}
    \end{subfigure}
    \begin{subfigure}[b]{0.24\columnwidth}
        \centering
         \includegraphics[width=\linewidth]{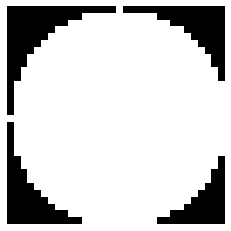}
    \end{subfigure}
    \caption{\textit{Stop}~\cite{zhong2022shadows} and \textit{Speed limit 30km/h} signs and the corresponding masks.}
    \label{fig:originalsignmask}
\end{figure}

To create an adversarial mask $M_A$, we initialize a noise tensor $N$ with the same dimensions as $I'$, filled with zeros. We optimize this noise over a series of time steps $t \in \{1, ..., T\}$, using gradients obtained from the model's output w.r.t. a given target class $c' \neq c$. The updates of the Adam optimizer~\cite{kingma2014adam} are restricted to an area $M_I$ that defines the region of the sign in the image $I'$. Otherwise, the noise could lie outside the traffic sign, leading to implausible results. Figure \ref{fig:originalsignmask} illustrates the stop sign and the corresponding mask $M_I$ by Zhong et al.~\cite{zhong2022shadows}. After $T$ iterations of gradient updates, we convert the final version of $N$ into a mask using an appropriate thresholding operation with the mask threshold $m$. 

\subsection{Phase II - Attack Generation}
In this phase, we aim to modify the pixels in the sign's region identified by the mask $M_A$ from the first phase. A new noise tensor $N$ is initialized with zeros, having the exact dimensions as $I'$. Using an optimizer, this noise tensor $N$ is updated over a series of time steps $t$. During each update, the noise is applied only to the region defined by the mask $M_A$, creating a perturbed image $I'_{\text{p}} = I' + N \cdot M_A$. This step is similar to a classical FGSM~\cite{goodfellow2014explaining} or PGD attack~\cite{madry2017towards}. To make the adversarial noise robust to the shift from the digital world to the real world, at each time step $t$, we apply random augmentations such as rotation, saturation, brightness, and translation to the perturbed image before we feed it into the model. This is inspired by the RP2 algorithm~\cite{eykholt2018robust}, however as we dispense the usage of physical transformations, it's more similar to an expectation over transformation (EOT) approach by Athalye et al.~\cite{athalye2018synthesizing}.

\subsection{Loss Function}
The loss function $L_{\text{total}}$ used in both phases comprises three terms: (1) a target class loss term, (2) a color consistency penalty term, and (3) a regularization term. The target class loss in the mask creation phase and training phase is the cross-entropy loss $L_{\text{ce}}(O, c'_T)$, where $O$ is the model's output, and $c'_T$ is the target class tensor.

\[ L_{\text{total}} = \beta_{1} \cdot L_{\text{color}}(N) + \beta_{2} \cdot L_{\text{reg1}}(N) + \beta_{3} \cdot L_{\text{ce}}(\Theta(I'), c'_T) \]
Where $\beta_{1}$, $\beta_{2}$, and $\beta_{3}$ are coefficients to adjust the influence of each term in the loss function.

The color consistency penalty term $L_{\text{color}}(N)$ is the same in both phases. It penalizes large differences between the color channels of the noise $N$ and is defined as:
\[ L_{\text{color}} = \sum_{i=1}^{C} \sum_{j=i+1}^{C} (N_{i} - N_{j})^2 \]
Where $C$ is the number of color channels (typically 3 for RGB images), with this loss term, it is possible to limit the attack to white, gray, or black colors, which should make it look like graffiti or dirt. Here, we try to improve the approach of Eykholt et al.~\cite{eykholt2018robust}. They use black and white stickers, which require manual human optimization~\cite{eykholt2018robust}; in our case, the white, gray, or black colors can be printed directly on the traffic sign.

The regularization term differs between phases. In the mask creation phase, the regularization term is the $l_1$ norm of the noise tensor $N$, $L_{\text{reg1}}(N) = \Vert N \Vert_1$. In the attack training phase, it is the $l_2$ norm of $N$, $L_{\text{reg2}}(N) = \Vert N \Vert_2$. Using the $l_1$ norm targets makes the noise more compact and less scattered over the sign. The $l_2$ restricts the insensitivity and amount of noise.

\subsection{Inconspicuous vs. Visible Attack}

We conduct our experiments using two distinct attack methodologies : (1) the \textit{inconspicuous attack} approach builds upon the techniques adapted from Eykholt et al.~\cite{eykholt2018robust} and Yang et al.~\cite{yang2021targeted}, enforcing inconspicuous attacks, whereas (2) the \textit{visible attack} approach mirrors the attack strategies presented by Chen et al.~\cite{chen2018shapeshifter} and Lu et al.~\cite{lu2017adversarial}, reflecting a visually similar but less regularized approach. For the latter, we have incorporated fewer regularization loss terms (see Table~\ref{tab:params}). 

\begin{table}[t]
	\centering
 	\caption{Attacks parameters.}
	\label{tab:params}
\begin{tabular}{|r|c|c|}
			\hline
			\textbf{Parameter}      & \textbf{Inconspicuous attack}         	 & \textbf{Visible attack}  \\
 \hline
            \rowcolor{LightBlue} \multicolumn{3}{|c|}{Phase I - Mask creation}  \\
            Number of epochs & 1000 & 100 \\
            Mask threshold $m$ & 0.1  & 0.05  \\
            $\beta_{1}$ & 0.1 & 0.1 \\
            $\beta_{2}$ & 0.01 & 0.001\\
            $\beta_{3}$ & 4.0 & 1.0 \\ \hline
            \rowcolor{LightBlue} \multicolumn{3}{|c|}{Phase II - Attack generation}  \\
            Number of epochs & 20K & 50K \\
            Learning rate & 0.01  & 0.0001  \\
            $\beta_{1}$ & 1.0 & 0.0\\
            $\beta_{2}$ & 0.0 & 0.0 \\
            $\beta_{3}$ & 4.0 & 1.0\\ 
            \hline
	\end{tabular}
\end{table}

\subsection{Evaluation}
To quantify the effectiveness of targeted adversarial attacks on our traffic sign classification algorithms, we define the attack success rate (ASR) as:

\begin{equation}
ASR_{targeted} = \frac{N_{successful\_targeted}}{N_{total}} \times 100,
\end{equation}

$N_{successful\_targeted}$ is the number of adversarial examples the model classifies as the specific incorrect class intended by the attacker, and $N_{total}$ is the total number of adversarial examples generated for testing. An ASR of 0 denotes complete resistance to the targeted adversarial attack. At the same time, an ASR of 100 means the model classified every adversarial input as the specific incorrect class intended by the attacker. Note that an ASR of 0 does not allow any conclusions to be drawn about the model's accuracy, which can also be 0.

\newpage
\section{Experiments and Evaluation}

\subsection{Experimental Setup}

\textbf{Datasets.} We employ two prominent traffic sign datasets in our experiments: the \texttt{LISA} Traffic Sign Dataset~\cite{mogelmose2012vision} and the German Traffic Sign Recognition Benchmark (GTSRB)~\cite{stallkamp2011gtsrb}.

The \texttt{LISA} dataset~\cite{mogelmose2012vision} encompasses a diverse set of over 7K annotated traffic sign instances from the USA; the dataset offers a wide variety of conditions, including day, night, and blur scenarios. The original collection features 47 different types of traffic signs. However, our work utilizes a subset of this dataset, specifically the version by Zhong et al.~\cite{zhong2022shadows}, comprising only the 16 most common classes. This version includes 6834 images, resized to $32\times32$, as is common in the literature~\cite{eykholt2018robust}. We use the 80:20 train-test split.

The GTSRB~\cite{stallkamp2011gtsrb} provides a comprehensive, multi-class, single-image classification challenge with 51839 images spanning 43 distinct classes of traffic signs. This includes a broad range of sign categories, such as speed limits, prohibitory signs, and danger signs. While the images in this dataset vary in size and are not consistently square, we apply the exact preprocessing step as with the \texttt{LISA} dataset and resize the images to a standard 32x32 pixels. The distribution of data for our experiments is split with the ratio 75.64:24.36.

\begin{figure}[tb]
    \centering
    \begin{subfigure}[t]{0.2\columnwidth}
        \centering
         \includegraphics[width=\linewidth]{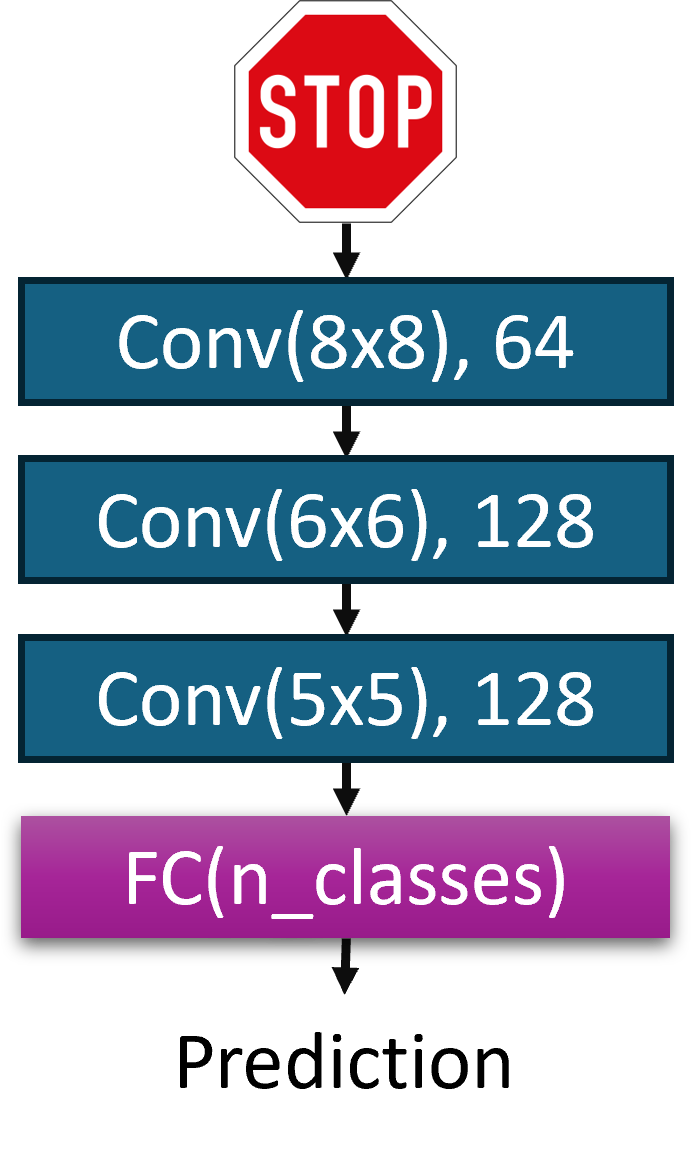}
         \caption{\texttt{CNN$_{small}$}}
    \end{subfigure}
    \hspace{15mm}
     \begin{subfigure}[t]{0.2\columnwidth}
        \centering
         \includegraphics[width=\linewidth]{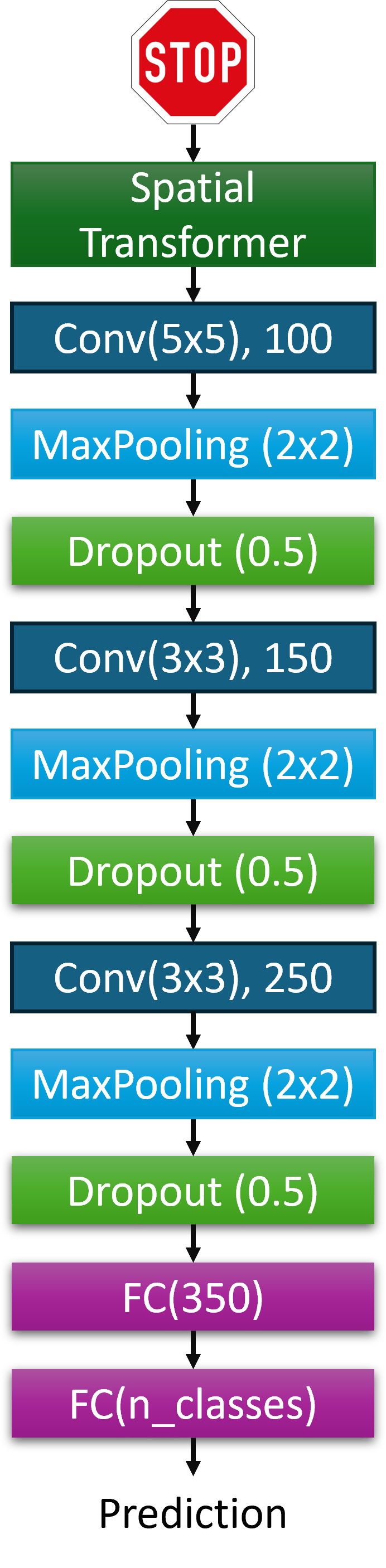}
         \caption{\texttt{CNN-STN} }
    \end{subfigure}
    \vfill
   \begin{subfigure}[t]{0.7\columnwidth}
        \centering
         \includegraphics[width=\linewidth]{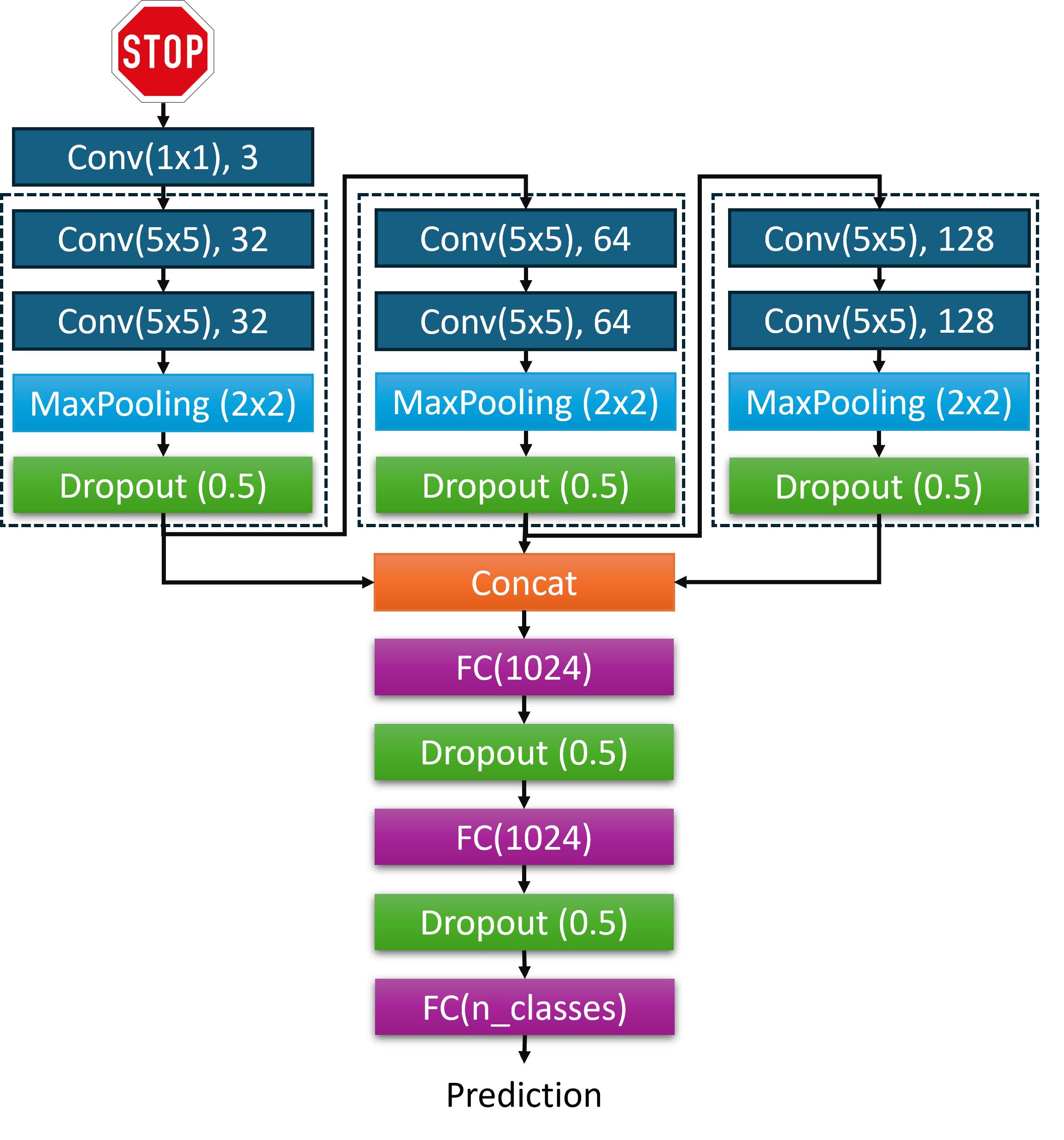}
         \caption{\texttt{CNN$_{large}$}}
    \end{subfigure}
   
    \caption{Baseline architectures. In \texttt{CNN$_{small}$} and  \texttt{CNN$_{large}$}, all \textit{Conv} layers use the \textit{ReLU} activation function. In \texttt{CNN-STN}, all \textit{Conv} layers use the \textit{LeakyReLU} activation function and batch normalization. The number of classes $n\_classes$ is 43 for \texttt{GTRSDB} and 16 for \texttt{LISA}.}
    \label{fig:architectures}
\end{figure}

\textbf{Models.} We consider three CNNs classically used for traffic sign classification (see Figure \ref{fig:architectures}):

\begin{enumerate}
    \item \texttt{CNN$_{small}$} is the original \texttt{LISA-CNN} as proposed by Eykholt et al.~\cite{eykholt2018robust}. We used the PyTorch implementation by Zhong et al.~\cite{zhong2022shadows}\footnote{\url{https://github.com/hncszyq/ShadowAttack}} and extended it to the \texttt{GTSRB} data. This is the smallest model with 739K parameters.
    \item \texttt{CNN$_{large}$} is the original \texttt{GTSRB-CNN} based on the multi-scale CNN~\cite{sermanet2011traffic} and a later implementation by Yadav\footnote{\url{https://github.com/vxy10/p2-TrafficSigns}}. We adapted the implementation from Zhong et al.~\cite{zhong2022shadows} and extended it to the \texttt{LISA} dataset. This model has the largest capacity with 16,571,223 parameters.
    \item \texttt{CNN-STN}: we used the original implementation by Garcia et al.~\cite{garcia2018deep}\footnote{\url{https://github.com/poojahira/gtsrb-pytorch}} and extended it to the \texttt{LISA} dataset. This model has 855,487 parameters.
\end{enumerate}

In addition to these architectures, deliberately developed for the traffic sign classification task, we have evaluated five further architectures, which are comparable or smaller in size: \texttt{ResNet18}~\cite{he2016deep}, \texttt{EfficientNet-B0}~\cite{tan2019efficientnet}, \texttt{DenseNet-121}~\cite{huang2017densely}, \texttt{MobileNetv2}~\cite{sandler2018mobilenetv2}, and \texttt{ShuffleNetv2} with 1.0x output~\cite{ma2018shufflenet}. We have decided against evaluating models like \texttt{VGG-16}~\cite{simonyan2014very}, \texttt{ResNet-34}, or \texttt{ResNet-50}, used in some previous works~\cite{wang2024traffic} because they significantly exceed the number of parameters in the three standard baselines mentioned above.

For training, we applied the same hyperparameters for all models: training for 100 epochs with a batch size of 64, using Adam optimizer~\cite{kingma2014adam} with a learning rate 0.01 and smooth cross-entropy loss with the smoothing factor 0.1.

\newpage

\subsection{Performance on Clean Data}
The best accuracy on both datasets was achieved with \texttt{CNN-STN} (see Table \ref{tab:baselineperformance}). For the smaller \texttt{LISA} dataset, comparable results were achieved with \texttt{ResNet18}. \texttt{CNN$_{large}$}, a standard baseline for \texttt{GTSRB} with the largest number of parameters, performed worse than \texttt{ResNet18}. On the other hand, \texttt{CNN$_{small}$}, which has so far been only used on \texttt{LISA} as \texttt{LISA-CNN},  performed well on \texttt{GTSRB}. \texttt{ShuffleNetV2} and \texttt{EfficientNet} have demonstrated the worst results on both datasets. In summary, \texttt{CNN-STN} is the only one out of three established baselines that outperform generic CNNs regarding accuracy and speed.

\begin{table}[t]
	\centering
 	\caption{Baseline performance on test data without attack.}
	\label{tab:baselineperformance}
        \begin{tabular}{|r|c|c|c|}
            \hline
            \textbf{Model} & \textbf{Accuracy, \%} & \textbf{Inf. speed, ms} & \textbf{\# parameters}\\ \hline
            \rowcolor{LightBlue} \multicolumn{4}{|c|}{\texttt{LISA} dataset}  \\
            \texttt{CNN$_{small}$} & 99.71 & 0.10409  & \textbf{0.73} M\\
            \texttt{CNN$_{large}$} & 99.78 & 0.05933   & 16.54 M\\
            \texttt{CNN-STN} & \textbf{99.85} & 0.13375   & 0.85 M\\ 
            \texttt{ResNet18} & \textbf{99.85} & 0.02589 & 11.18 M\\
            \texttt{MobileNetv2} & 99.71  & \textbf{0.05265}   & 3.50 M\\
            \texttt{DenseNet} & 99.63  & 0.16382   &  7.98 M\\
            \texttt{ShuffleNetV2} & 99.27 & 0.08542 & 2.28 M\\
            \texttt{EfficientNet} & 99.34 & 0.08185  & 5.29 M\\ \hline 
            
            \rowcolor{LightBlue} \multicolumn{4}{|c|}{\texttt{GTSRB} dataset}  \\
            \texttt{CNN$_{small}$} & 98.13 & 0.00651   & \textbf{0.74} M\\
            \texttt{CNN$_{large}$} & 98.91 & 0.00972   & 16.57 M\\
            \texttt{CNN-STN} & \textbf{99.43} & 0.02231   & 0.86 M\\
            \texttt{ResNet18} & 99.18 & 0.02511  & 11.19 M\\
            \texttt{MobileNetv2} & 98.36  & 0.0475   & 3.50 M\\
            \texttt{DenseNet} & 98.09 & 0.1287  & 7.98 M\\
            \texttt{ShuffleNetV2} & 96.06  & 0.06014  &  2.28 M\\
             \texttt{EfficientNet} & 98.73  & 0.07186   & 5.29 M\\
            \hline
        \end{tabular}
\end{table}

\begin{table}[t]
	\centering
 	\caption{Predicted class and the corresponding confidence under digital inconspicuous and visible attacks.}
	\label{tab:attackresults}
    		\begin{tabular}{|r|l  |l  |l  |}
    			\hline
    			\textbf{Model}  & \textbf{No attack} & \textbf{Inconspicuous} & \textbf{Visible} \\
                    &  & \textbf{attack} & \textbf{attack} \\\hline
 \rowcolor{LightBlue} \multicolumn{4}{|c|}{\texttt{LISA} dataset, \includegraphics[width=0.015\textwidth]{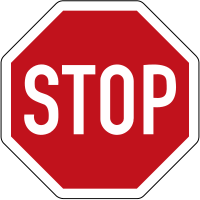} $\rightarrow$ \includegraphics[width=0.015\textwidth]{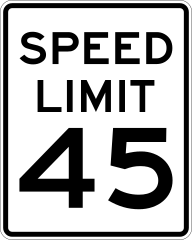} attack.} \\
 
        \texttt{CNN$_{small}$} & \includegraphics[width=0.015\textwidth]{img/stopsign.png}  78.32 & \includegraphics[width=0.015\textwidth]{img/Speed_Limit_45_sign.svg.png}  \textbf{95.35}     & \includegraphics[width=0.015\textwidth]{img/Speed_Limit_45_sign.svg.png}  99.31  \\

        \texttt{CNN$_{large}$}   & \includegraphics[width=0.015\textwidth]{img/stopsign.png}  \textbf{92.66}& \includegraphics[width=0.015\textwidth]{img/Speed_Limit_45_sign.svg.png}  79.66    & \includegraphics[width=0.015\textwidth]{img/Speed_Limit_45_sign.svg.png} 95.56  \\
       
         \texttt{CNN-STN}  & \includegraphics[width=0.015\textwidth]{img/stopsign.png}  91.03 & \includegraphics[width=0.015\textwidth]{img/Speed_Limit_45_sign.svg.png} 88.64     & \includegraphics[width=0.015\textwidth]{img/Speed_Limit_45_sign.svg.png} \textbf{99.99}  \\
        \texttt{ResNet18}  & \includegraphics[width=0.015\textwidth]{img/stopsign.png}  89.82 & \includegraphics[width=0.015\textwidth]{img/Speed_Limit_45_sign.svg.png} 30.88    & \includegraphics[width=0.015\textwidth]{img/Speed_Limit_45_sign.svg.png}  89.37  \\
         \texttt{MobileNetv2}  & \includegraphics[width=0.015\textwidth]{img/stopsign.png}  87.37 & \includegraphics[width=0.015\textwidth]{img/stopsign.png} 86.35   & \includegraphics[width=0.015\textwidth]{img/stopsign.png}  86.87  \\
         \texttt{DenseNet}  & \includegraphics[width=0.015\textwidth]{img/stopsign.png}  78.45 & \includegraphics[width=0.015\textwidth]{img/Speed_Limit_45_sign.svg.png} 35.41     & \includegraphics[width=0.015\textwidth]{img/Speed_Limit_45_sign.svg.png} 90.30  \\
         \texttt{ShuffleNetV2}  & \includegraphics[width=0.015\textwidth]{img/stopsign.png}  90.01 & \includegraphics[width=0.015\textwidth]{img/stopsign.png} 90.08   & \includegraphics[width=0.015\textwidth]{img/stopsign.png} 78.81  \\
         \texttt{EfficientNet}  & \includegraphics[width=0.015\textwidth]{img/stopsign.png} 88.41 & \includegraphics[width=0.015\textwidth]{img/Speed_Limit_45_sign.svg.png} 47.02    & \includegraphics[width=0.015\textwidth]{img/stopsign.png} 44.48 \\ \hline
 
  \rowcolor{LightBlue} \multicolumn{4}{|c|}{\texttt{LISA} dataset, \includegraphics[width=0.015\textwidth]{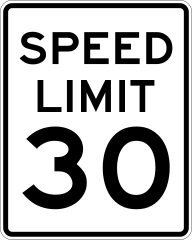} $\rightarrow$ \includegraphics[width=0.015\textwidth]{img/stopsign.png} attack.}  \\
 
        \texttt{CNN$_{small}$} & \includegraphics[width=0.015\textwidth]{img/Speed_Limit_30_sign.svg.png} \textbf{95.07} & \includegraphics[width=0.015\textwidth]{img/stopsign.png}   \textbf{79.55} & \includegraphics[width=0.015\textwidth]{img/stopsign.png}  \textbf{97.11}    \\

        \texttt{CNN$_{large}$}   & \includegraphics[width=0.015\textwidth]{img/Speed_Limit_30_sign.svg.png} 86.76 & \includegraphics[width=0.015\textwidth]{img/stopsign.png} 42.95 & \includegraphics[width=0.015\textwidth]{img/stopsign.png}  92.84  \\
       
         \texttt{CNN-STN}  & \includegraphics[width=0.015\textwidth]{img/Speed_Limit_30_sign.svg.png} 94.22  & \includegraphics[width=0.015\textwidth]{img/stopsign.png}  36.64     & \includegraphics[width=0.015\textwidth]{img/stopsign.png}  91.29 \\
        \texttt{ResNet18}  & \includegraphics[width=0.015\textwidth]{img/Speed_Limit_30_sign.svg.png} 87.31 & \includegraphics[width=0.015\textwidth]{img/stopsign.png}  33.73  & \includegraphics[width=0.015\textwidth]{img/stopsign.png} 70.33  \\
         \texttt{MobileNetv2}  & \includegraphics[width=0.015\textwidth]{img/Speed_Limit_30_sign.svg.png} 90.33 & \includegraphics[width=0.015\textwidth]{img/Speed_Limit_30_sign.svg.png} 90.52   & \includegraphics[width=0.015\textwidth]{img/Speed_Limit_30_sign.svg.png} 52.19    \\
         \texttt{DenseNet}  & \includegraphics[width=0.015\textwidth]{img/Speed_Limit_30_sign.svg.png} 86.97 & \includegraphics[width=0.015\textwidth]{img/Speed_Limit_30_sign.svg.png} 79.21      & \includegraphics[width=0.015\textwidth]{img/stopsign.png}  83.06  \\
         \texttt{ShuffleNetV2}  & \includegraphics[width=0.015\textwidth]{img/Speed_Limit_30_sign.svg.png} 88.38 & \includegraphics[width=0.015\textwidth]{img/Speed_Limit_30_sign.svg.png} 83.09   & \includegraphics[width=0.015\textwidth]{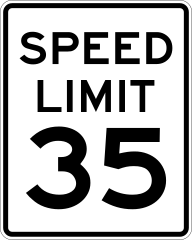}  35.30 \\
         \texttt{EfficientNet}  & \includegraphics[width=0.015\textwidth]{img/Speed_Limit_30_sign.svg.png} 62.22 & \includegraphics[width=0.015\textwidth]{img/Speed_Limit_30_sign.svg.png} 59.89  & \includegraphics[width=0.015\textwidth]{img/Speed_Limit_30_sign.svg.png}  51.67\\ \hline
 
 \rowcolor{LightBlue} \multicolumn{4}{|c|}{\texttt{GTSRB} dataset, \includegraphics[width=0.015\textwidth]{img/stopsign.png} $\rightarrow$ \includegraphics[width=0.015\textwidth]{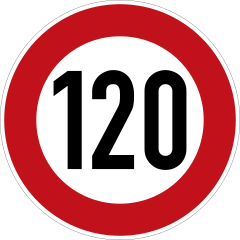} attack.}  \\
         \texttt{CNN$_{small}$}   & \includegraphics[width=0.015\textwidth]{img/stopsign.png}  \textbf{87.16}              & \includegraphics[width=0.015\textwidth]{img/120kmh.png}  20.50 & \includegraphics[width=0.015\textwidth]{img/120kmh.png}   90.18  \\  
         \texttt{CNN$_{large}$}    &  \includegraphics[width=0.015\textwidth]{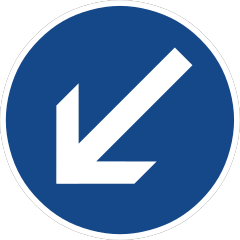} 8.18   & \includegraphics[width=0.015\textwidth]{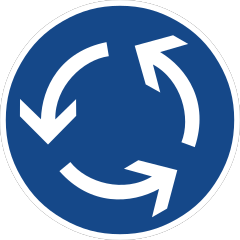}  12.87  & \includegraphics[width=0.015\textwidth]{img/120kmh.png}  94.66  \\
         \texttt{CNN-STN}    & \includegraphics[width=0.015\textwidth]{img/stopsign.png} 72.85            & \includegraphics[width=0.015\textwidth]{img/120kmh.png}    \textbf{52.14}  & \includegraphics[width=0.015\textwidth]{img/120kmh.png}  \textbf{99.09}  \\            
        \texttt{ResNet18}  & \includegraphics[width=0.015\textwidth]{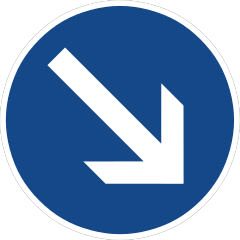}  72.58 & \includegraphics[width=0.015\textwidth]{img/keepright.png}  37.95  & \includegraphics[width=0.015\textwidth]{img/120kmh.png}  97.96 \\
        \texttt{MobileNetv2}  & \includegraphics[width=0.015\textwidth]{img/roundabout.png}  37.04  & \includegraphics[width=0.015\textwidth]{img/roundabout.png}   44.48 & \includegraphics[width=0.015\textwidth]{img/120kmh.png} 93.27  \\
        \texttt{DenseNet}    & \includegraphics[width=0.015\textwidth]{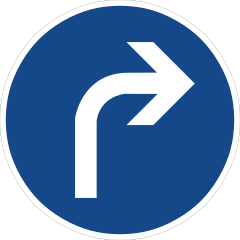}  28.01 & \includegraphics[width=0.015\textwidth]{img/turn_right.png} 33.23    & \includegraphics[width=0.015\textwidth]{img/120kmh.png} 93.07 \\
        \texttt{ShuffleNetV2}    & \includegraphics[width=0.015\textwidth]{img/roundabout.png}  57.33 & \includegraphics[width=0.015\textwidth]{img/roundabout.png} 27.56   & \includegraphics[width=0.015\textwidth]{img/120kmh.png} 96.82\\
        \texttt{EfficientNet}    & \includegraphics[width=0.015\textwidth]{img/stopsign.png}  84.00   & \includegraphics[width=0.015\textwidth]{img/120kmh.png}  38.65  & \includegraphics[width=0.015\textwidth]{img/120kmh.png}   93.54 \\\hline

 \rowcolor{LightBlue} \multicolumn{4}{|c|}{\texttt{GTSRB} dataset, \includegraphics[width=0.015\textwidth]{img/30kmh.png} $\rightarrow$ \includegraphics[width=0.015\textwidth]{img/stopsign.png} attack.}  \\
     \texttt{CNN$_{small}$}  & \includegraphics[width=0.015\textwidth]{img/30kmh.png}  88.50      & \includegraphics[width=0.015\textwidth]{img/30kmh.png} 31.44 & \includegraphics[width=0.015\textwidth]{img/stopsign.png}  92.96\\  
     \texttt{CNN$_{large}$}    &  \includegraphics[width=0.015\textwidth]{img/30kmh.png} \textbf{90.89}  & \includegraphics[width=0.015\textwidth]{img/30kmh.png} \textbf{90.89}  & \includegraphics[width=0.015\textwidth]{img/stopsign.png} 97.87 \\
     \texttt{CNN-STN}    & \includegraphics[width=0.015\textwidth]{img/30kmh.png}  90.40 & \includegraphics[width=0.015\textwidth]{img/30kmh.png}  57.61 & \includegraphics[width=0.015\textwidth]{img/stopsign.png} 98.10 \\            
    \texttt{ResNet18}  & \includegraphics[width=0.015\textwidth]{img/30kmh.png} 65.99 & \includegraphics[width=0.015\textwidth]{img/30kmh.png} 6.41 & \includegraphics[width=0.015\textwidth]{img/stopsign.png} 88.79\\
    \texttt{MobileNetv2}  & \includegraphics[width=0.015\textwidth]{img/30kmh.png} 84.87 & \includegraphics[width=0.015\textwidth]{img/30kmh.png} 26.32 & \includegraphics[width=0.015\textwidth]{img/stopsign.png} 75.81 \\
    \texttt{DenseNet}    & \includegraphics[width=0.015\textwidth]{img/30kmh.png} 77.92 &  \includegraphics[width=0.015\textwidth]{img/30kmh.png} 24.77 & \includegraphics[width=0.015\textwidth]{img/stopsign.png} 93.84\\
    \texttt{ShuffleNetV2}    & \includegraphics[width=0.015\textwidth]{img/keepright.png} 97.39 & \includegraphics[width=0.015\textwidth]{img/keepright.png} 96.77  & \includegraphics[width=0.015\textwidth]{img/stopsign.png}  65.47\\
    \texttt{EfficientNet}    & \includegraphics[width=0.015\textwidth]{img/30kmh.png} 76.70  & \includegraphics[width=0.015\textwidth]{img/120kmh.png} 25.85 & \includegraphics[width=0.015\textwidth]{img/stopsign.png} 87.03 \\\hline

    	\end{tabular}
\end{table}

\newpage
\clearpage

\begin{figure}[t!]
    \centering
    \resizebox{1.0\linewidth}{!}{
    \begin{tabular}{c c c c c}
       
        \texttt{CNN$_{small}$} & \texttt{CNN$_{large}$} &  \texttt{CNN-STN} &  \texttt{ResNet18}  & \texttt{EfficientNet}\\
        \includegraphics[width=0.09\textwidth]{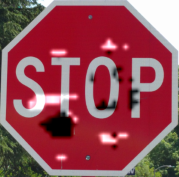} &
        \includegraphics[width=0.09\textwidth]{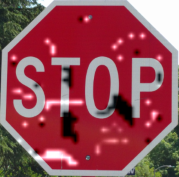} &
        \includegraphics[width=0.09\textwidth]{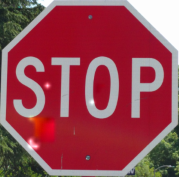}&
        \includegraphics[width=0.09\textwidth]{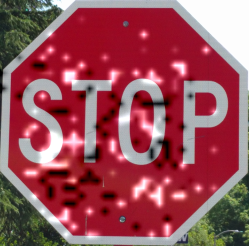}& 
        \includegraphics[width=0.09\textwidth]{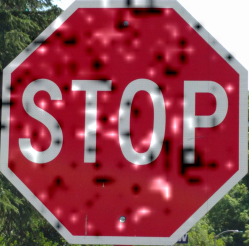}
        \\
        \multicolumn{5}{c}{a) \texttt{LISA} dataset - inconspicuous attack}  \\

        \includegraphics[width=0.09\textwidth]{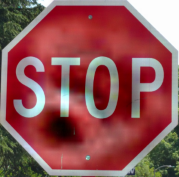} & 
        \includegraphics[width=0.09\textwidth]{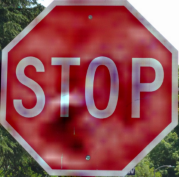} & 
        \includegraphics[width=0.09\textwidth]{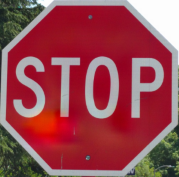} & 
        \includegraphics[width=0.09\textwidth]{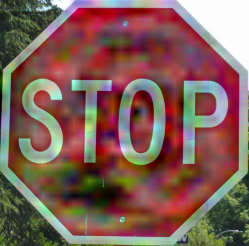} & 
        \includegraphics[width=0.09\textwidth]{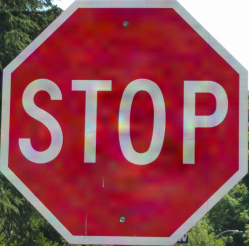}\\
         \multicolumn{5}{c}{b) \texttt{LISA} dataset - visible attack}  \\

        \includegraphics[width=0.09\textwidth]{img/CNNsmallGTSRBinconspicuous1.png} &
        \includegraphics[width=0.09\textwidth]{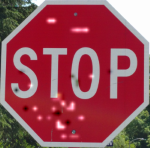} &
        \includegraphics[width=0.09\textwidth]{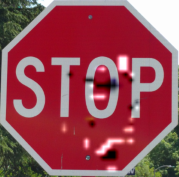} & 
        \includegraphics[width=0.09\textwidth]{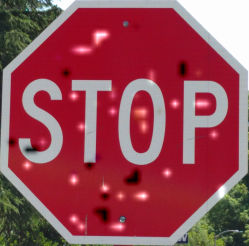} & 

        \includegraphics[width=0.09\textwidth]{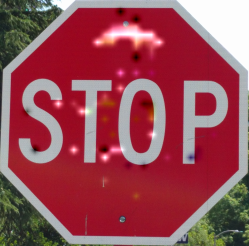}\\
        \multicolumn{5}{c}{c) \texttt{GTSRB} dataset - inconspicuous attack}  \\

        \includegraphics[width=0.09\textwidth]{img/CNNsmallGTSRBoffensive1.png} &
        \includegraphics[width=0.09\textwidth]{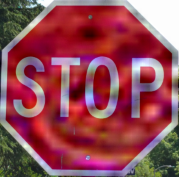} &
        \includegraphics[width=0.09\textwidth]{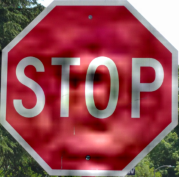} & 
        \includegraphics[width=0.09\textwidth]{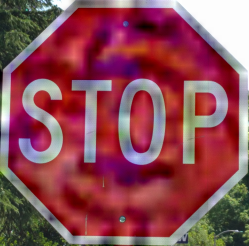}& 
        \includegraphics[width=0.09\textwidth]{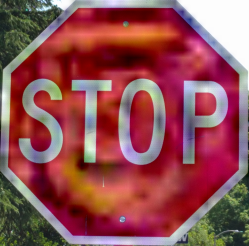}\\
        \multicolumn{5}{c}{d) \texttt{GTSRB} dataset - visible attack}  \\
        
    \end{tabular}
    }
    \caption{Examples of \textit{inconspicuous} and \textit{visible} attacks on a \textit{Stop} sign.}
    \label{fig:adversarialsigns}
\end{figure}

\subsection{Evaluation in the Digital Domain}

Our experiments in the digital domain with attacks on a \textit{Stop} sign and on a \textit{Speed limit 30} sign demonstrate significant variance in the way models react to the attacks (see Table~\ref{tab:attackresults}). \texttt{GTSRB}-based models tend to misclassify even clean data, and the attacks are less successful than on \texttt{LISA}-based models. This can be explained by the greater complexity of the \texttt{GTSRB} dataset regarding the number of instances and classes. 

Of all the evaluated models, \texttt{CNN$_{small}$} models are the easiest to attack and demonstrate the highest confidence in the target class. \texttt{MobileNetv2} and \texttt{ShuffleNetV2} are resistant to both inconspicuous and visible attacks. Also, other generic CNNs demonstrate lower confidence in the target class for both types of attacks. Of all generic models, \texttt{ResNet18} and \texttt{EfficientNet} were prone to the studied attacks. Generally, generic models are much less vulnerable to studied attacks, although they demonstrate lower accuracy on clean data. Especially for the \texttt{GTSRB} data, attacks on established baselines cannot be reproduced on generic ones. These results stress that new attacks should be evaluated beyond standard baselines.

The appearance of the resulting perturbations also varies a lot (see Figure~\ref{fig:adversarialsigns}). While for some models, the l1 norm in the inconspicuous attack induces sticker-like permutations (as with \texttt{CNN$_{small}$}), others develop negligible noise for the same attack (such as \texttt{CNN-STN}). We observe similar results for the visible attacks. For instance, the \texttt{CNN$_{small}$} sign is vibrant, reminiscent of the work by Yang et al.~\cite{yang2021targeted}, whereas the sign for \texttt{CNN-STN} gives the impression of having used the color penalty term. However, $\beta_3=0$ for both signs. Another exciting discovery involves the results for \texttt{CNN-STN} for \texttt{LISA}. The signs resemble the original ones closely, yet the model predicts them as speed limit with a confidence of over $99\%$ for both signs.

\subsection{Real-World Evaluation}

We perform a two-fold evaluation in the real world to assess and compare the effectiveness of the various attack strategies. First, we print each traffic sign in its manipulated form. Then, we simulate a \textit{drive-through} scenario for each sign to replicate the real-world conditions under which these signs would typically be viewed. We pinned images on a white wall and moved the camera from far away, directly "through" the traffic sign. Additionally, we compare a video with an unaltered drive-by of an original \textit{Stop} sign to provide a baseline.

Table \ref{tab:drivethroughresults} presents the results of our \textit{drive-through} experiments. We can observe the performance of the six baseline models on three signs (original, inconspicuous, and visible), expressed as the attack success rate. The attack success rate on the original sign is $0$ for all models, i.e., none of the models misclassify the original \textit{Stop} sign as the target class. However, the performance of the models on adversarial signs varies. Generally, the models trained on the \texttt{LISA} dataset seem more susceptible to attack, which we believe is due to the smaller dataset size. Another observation is the higher ASR of visible attacks than inconspicuous ones.

Furthermore, \texttt{CNN-STN} models were successfully attacked during training, as seen in Table \ref{tab:attackresults}. However, they only worked in three out of four real-world cases (see Table \ref{tab:drivethroughresults}). Figure \ref{fig:adversarialsigns} shows that the perturbations are almost invisible. This could be the reason for poor results during the \textit{drive-through} evaluation but does not explain the discrepancy between the digital and real setting.

\begin{table}[t]
	\centering
 	\caption{Attack success rates for the \textit{drive-through} attacks on the perturbed \textit{Stop} signs.}
	\label{tab:drivethroughresults}
		\begin{tabular}{|r|c|c|}
			\hline
			\textbf{Baseline model}      & \textbf{Inconspicuous attack}         	 & \textbf{Visible attack}  \\
 \hline
			\rowcolor{LightBlue} \multicolumn{3}{|c|}{\texttt{LISA} dataset}  \\
            \texttt{CNN$_{small}$}        & \textbf{100.0\%}   & \textbf{100.0\%}\\
			\texttt{CNN$_{large}$}              & 52.01\%      & \textbf{61.0\%} \\
            \texttt{CNN-STN}           & 0.0\%         & 0.0\%  \\
            \hline
            
            \rowcolor{LightBlue} \multicolumn{3}{|c|}{\texttt{GTSRB} dataset}  \\
\texttt{CNN$_{small}$}& 1.71\%        & \textbf{39.14\%}\\ 
 \texttt{CNN$_{large}$} & 2.71\%        & \textbf{4.62\%} \\
 \texttt{CNN-STN}             & 0.0\%         & \textbf{100.0\%}\\
            \hline
	\end{tabular}
\end{table}

\section{Conclusion}
\label{conclusionandprospects}

In this work, we experimentally compared adversarial attacks on traffic sign classification tasks across multiple model architectures and datasets in inconspicuous and visible settings. We have decoupled three established architectures from the datasets they are traditionally trained on and also evaluated five further generic image classification CNNs. Our experiments in the digital and physical domains have shown that three established baselines are more susceptible to attacks than the generic ones. Based on our findings, we suggest adapting the evaluation protocol for adversarial attacks on traffic sign classification models and particularly evaluating new attacks on a broader range of models, including more robust generic image classification models.

\newpage
\section*{Acknowledgment}

This work was supported by funding from the Topic Engineering Secure Systems of the Helmholtz Association (HGF) and by KASTEL Security Research Labs (46.23.03).

{\small
\bibliographystyle{IEEEtran}
\bibliography{references}
}

\end{document}